\documentclass[letterpaper, 10pt, journal, twoside]{IEEEtran}  % Comment this line out if you need a4paper

\IEEEoverridecommandlockouts                              % This command is only needed if 
\usepackage{amsmath,amsfonts,amssymb}
\usepackage{array}
\usepackage[
    caption=false,
    font=normalsize,
    labelfont=sf,
    textfont=sf
]{subfig}
\usepackage{textcomp}
\usepackage{verbatim}
\usepackage{graphicx}
\usepackage{cite}
\usepackage{siunitx}
\usepackage{mathrsfs}
\usepackage{bm}
\usepackage{multicol}
\usepackage{multirow}
\usepackage{makecell}
\usepackage[dvipsnames]{xcolor}

\usepackage[inkscapelatex=false]{svg}
\usepackage{adjustbox}

\usepackage{algorithm}
\usepackage{algpseudocode}

\usepackage{comment}

\usepackage[
    bookmarks=false,
    colorlinks=true,
    linkcolor=black,
    citecolor=black,
    urlcolor=black
]{hyperref}

\title{ 
Differentiable Dynamics and Fast Simulation of Continuous Elastic Robotic Fish
}
\author{Zhiheng Chen$^{1,2}$, Jiayi Jin$^{1}$, and Wei Wang$^1$
\thanks{$^{1}$Marine Robotics Lab, Department of Mechanical Engineering, College of Engineering, University of Wisconsin-Madison.}
\thanks{$^{2}$Department of Mechanical Engineering, University of Michigan.}
\thanks{
$^{\ast}$Corresponding author: {\tt\small wwang745@wisc.edu}.}
}

\begin{document}

% % Paper headers
% \markboth{IEEE Robotics and Automation Letters.}
% { \MakeLowercase{\textit{et al.}}:
% %Differentiable Modeling and Fast Simulation of
% %Continuous Elastic-Body Robotic Fish
% } 
% % Use only for final RAL version

\IEEEpubid{ }

\maketitle
\thispagestyle{empty}

% \pagestyle{empty}

%%%%%%%%%%%%%%%%%%%%%%%%%%%%%%%%%%%%%%%%%%%%%%%%%%%%%%%%%%%%%%%%%%%%%%%%%%%%%%%%
\begin{abstract}
%Fish-inspired aquatic robots are gaining increasing attention in marine robot communities due to their high swimming speeds and efficient propulsion enabled by flexible bodies that generate undulatory motions. To support the design optimization and control of such systems, accurate, interpretable, and computationally tractable modeling of the underlying swimming dynamics is indispensable. In this letter, we present a differentiable full-body dynamics model and a fast simulator for motor-actuated robotic fish, rigorously derived from Hamilton’s principle. The model captures the continuously distributed elasticity of a deformable fish body undergoing large deformations and incorporates fluid–structure coupling effects, enabling self-propelled motion without prescribing kinematics. Numerical convergence and hardware experiments validate the dynamics model, and the differentiability of the simulation framework is demonstrated by numerical optimizations over the body stiffness distribution.

Body flexibility plays a critical role in fish-like swimming, as the spatial distribution of stiffness governs body deformation, hydrodynamic loading, and propulsive performance. Exploiting this mechanism in robotic fish requires dynamic models that capture continuous body elasticity, fluid–structure interaction, and the resulting self-propelled motion. Existing approaches often prescribe body kinematics, approximate the body using discrete rigid or compliant segments, or incur high computational costs that limit their use in design optimization. In this letter, we present a differentiable full-body dynamics model and fast simulation framework for motor-actuated elastic robotic fish based on Hamilton’s principle. The proposed formulation represents the robot as a continuously deformable elastic body and couples its structural dynamics with hydrodynamic forces without prescribing body kinematics. The resulting simulator is differentiable with respect to model and design parameters, enabling efficient gradient-based optimization. Numerical convergence studies and experiments with a physical robotic fish validate the proposed framework. Finally, gradient-based optimization of the body stiffness distribution demonstrates its utility for efficient design of elastic robotic fish.

% Preliminary open-loop simulations examine how actuation frequency and body stiffness influence the swimming speed and energy efficiency of the robotic fish. Closed-loop simulations further assess how stiffness distribution impacts the controller’s velocity-tracking performance and energy efficiency. The results demonstrate the model’s potential for performance evaluation and control optimization of soft robotic swimmers when stiffness is treated as a design variable.

% Results reveal distinct trade-offs between propulsion speed and cost of transport (COT), demonstrating the model’s capability to support comprehensive performance evaluations and control analyses for soft robotic swimmers.
\end{abstract}

% \begin{IEEEkeywords}
% % List of keywords (from the RA Letters keyword list)
% Robotic fish, elastic-body dynamics, differentiable simulation, stiffness optimization.
% \end{IEEEkeywords}

%%%%%%%%%%%%%%%%%%%%%%%%%%%%%%%%%%%%%%%%%%%%%%%%%%%%%%%%%%%%%%%%%%%%%%%%%%%%%%%%
\section{Introduction}

Biomimetic swimming robots have attracted growing interest in marine robotics because of their potential for energy-efficient and agile locomotion in complex aquatic environments \cite{li2023,liu2025}. Among them, fish-like robots are particularly promising, as they exploit body flexibility and undulatory motion to achieve high swimming speeds, propulsive efficiency, and maneuverability \cite{iguchi2024,van2022,quinn2021}. Because their performance is governed by the coupled interaction between body deformation and hydrodynamic forces, accurate dynamics modeling is essential not only for reliable simulation and control, but also for design optimization and performance enhancement of bio-inspired robotic swimmers.

Considerable theoretical and computational effort has been devoted to modeling the dynamics of robotic fish. A widely adopted approach is the multi-segment formulation \cite{li2014modeling, zheng2022three,wang2015,liao2022,jiang2022,mcmillen2006}, in which the fish body is represented as a chain of interconnected segments. In most of these models, the segments are treated as rigid bodies, while body elasticity is lumped at the joints through torsional springs. Although \cite{mcmillen2006} is derived from an elastic-rod formulation, its numerical implementation similarly discretizes the body into rigid segments, with bending elasticity represented by local potentials at the segment connections. A notable exception is \cite{liao2022}, which approximates each segment as a deformable arc, introducing an additional degree of freedom to represent within-segment bending.
Although multi-segment methods can be computationally efficient when only a few segments are used, they inherently approximate the continuous deformation of a fish body using discrete degrees of freedom. In real fish, body curvature varies smoothly along the body rather than changing abruptly at isolated joint locations. Consequently, coarse spatial discretization can introduce non-negligible errors in representing traveling body waves and the resulting fluid–structure interactions. 
%Such errors may limit the fidelity of propulsion analysis, for which accurate representation of smooth, continuous undulatory motion is essential.

Another approach to modeling robotic fish dynamics is through numerical simulations based on computational fluid dynamics (CFD) and fluid–structure interaction (FSI). The FSI framework in \cite{curatolo2016} provides high-fidelity simulations of muscle-actuated fish swimming. Similarly, \cite{sohn2021} presents a CFD model that accurately resolves vortex shedding during swimming; however, the body deformation is prescribed, and therefore the model does not capture the fully coupled body dynamics. The FSI model in \cite{lu2025} provides detailed analysis of the surrounding flow field and the influence of stiffness tuning on propulsive performance. However, its structural model is simplified to two rigid bodies connected by a joint with a nonlinear spring and damper, limiting its ability to represent continuous body deformation.
Despite their potentially high fidelity, CFD- and FSI-based models generally do not yield compact, differentiable, and interpretable closed-form equations of motion, which are highly desirable for robot design optimization  and model-based control. Moreover, high-fidelity CFD and FSI simulations are computationally intensive \cite{o2021,Wu_2021}, which makes iterative design optimization and real-time model-based control impractical.

Beam-theoretic approaches have also been used to study fish swimming dynamics. In \cite{cheng1998}, an Euler–Bernoulli beam model is developed, but it assumes small body deformations and prescribes a constant forward swimming speed, both of which limit its applicability to realistic self-propelled swimming. In \cite{kopman2012}, the fishtail is modeled as a hinged beam and its linear vibration modes are analyzed. This formulation likewise assumes small deformations, and by confining flexibility to the tail, it cannot readily represent swimming modes such as anguilliform or subcarangiform locomotion.

Motivated by these limitations, this letter develops  differentiable full-body dynamics model and fast simulation framework for elastic robotic
fish swimming. The model represents body elasticity as a continuous spatial distribution, accommodates arbitrarily large deformations, and incorporates both reactive and resistive hydrodynamic forces. Self-propelled motion emerges directly from the coupled dynamics without imposing prescribed body kinematics. We validate the model through numerical convergence studies and experiments with a physical robotic fish, and further demonstrate the differentiability of the simulation framework through gradient-based optimization of the body stiffness distribution to balance swimming speed and efficiency.
% , investigating swimming performance with varying actuation frequencies and body stiffness, and in the presence of closed-loop control, thereby providing insights into both the fundamental dynamics and the design of efficient robotic swimmers, as well as how stiffness distribution can influence the performances of a feedback controller, especially how a small change in the baseline elasticity can considerably alter the cost of transport at different target speeds.

\section{Kinematical Overview}
\label{sec: kinematics}
\thispagestyle{empty}

\begin{figure}[!htb]
    \centering
    \includegraphics[width=0.95\linewidth]{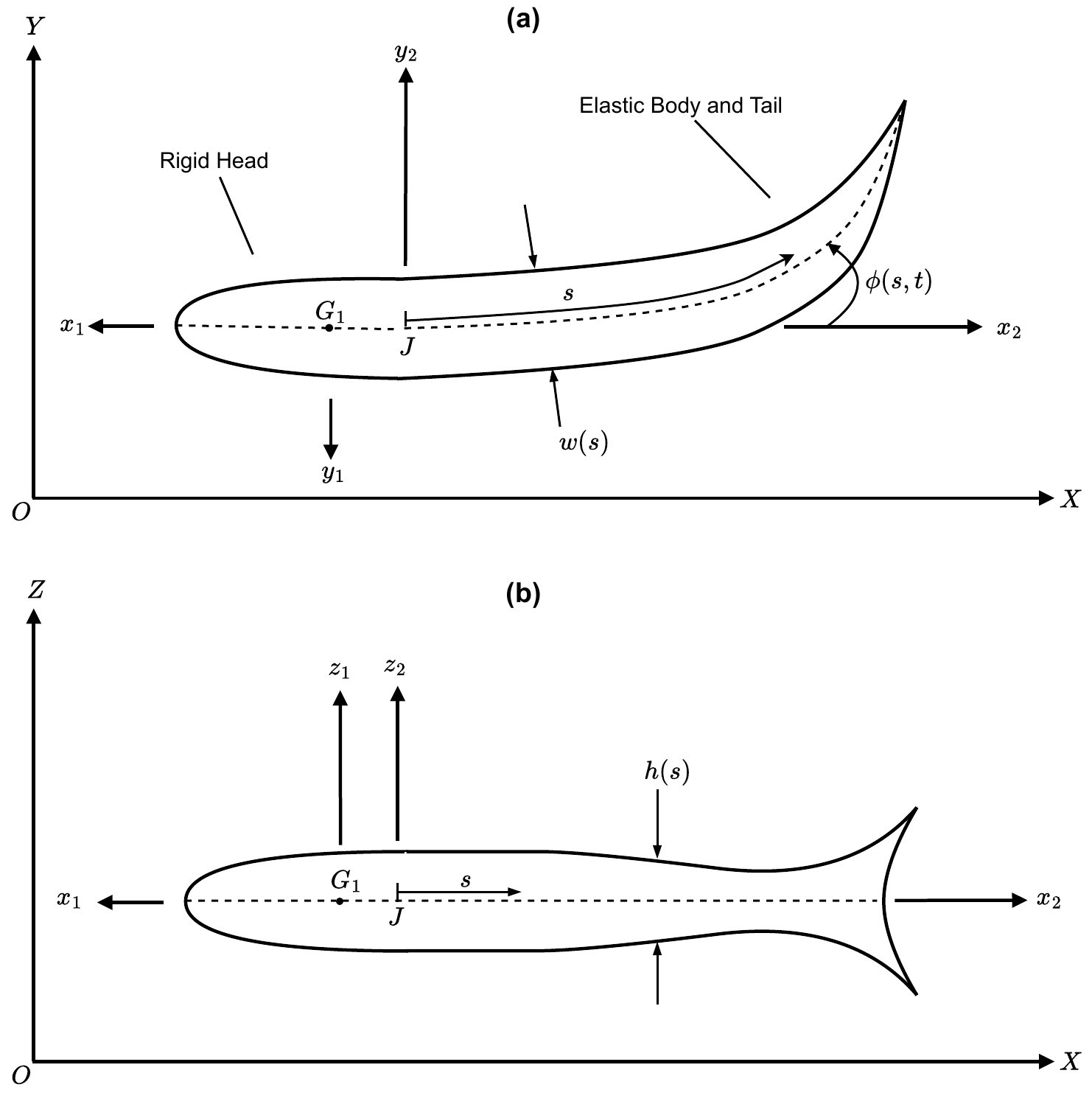}
    \caption{Configuration and coordinate systems of the fish model; $J$ stands for the revolute joint between the head and the body. (a) Top view and (b) side view of the fish.}
    \label{fig: configuration}
\end{figure}

As shown in Fig. \ref{fig: configuration}, we model the robotic fish as a rigid head and an elastic body connected by a motor-driven revolute joint \cite{van2022}; this modeling scheme is also capable of capturing many other types of structures, such as the design where a wire-driven plate is placed at the connection between the head and the body, with the driving motor attached to the head \cite{mazumdar2008}.  

Let $XYZ$ be the inertial frame, attach $x_1y_1z_1$ to the center of mass (COM) $G_1$ of the fish's head, and attach $x_2y_2z_2$ to the joint connecting the head and the body, so that it rotates with the head. We model the cross-section of the fish body as an ellipse whose axis lengths $w(s)$ and $h(s)$ can vary along the body.

\subsection{Generalized Coordinates and Displacement Fields}
We focus on the planar maneuvers of the fish. Thus, we model the head of the fish as a rigid body described by 3 unconstrained generalized coordinates: $X_{G_1}$, $Y_{G_1}$, and $\theta_{G_1}$. We model the fish's body as Euler's elastica \cite{caflisch1984} that can go through large deformations in the $x_2y_2$-plane. The tangential angle $\phi(s,t)$ describes the fish body's relative displacement field, where $s$ denotes the Lagrangian coordinate along the fish body. Under this framework, we describe the rotation of the fish's body about the joint as a special deformation represented by the rigid-body mode (more detailed discussions in Section \ref{sec: Ritz and computations}). To facilitate the following derivations, we also utilize the coordinates $x_2(s,t)$ and $y_2(s,t)$ for each point along the fish body. The coordinates are related to the displacement field $\phi(s,t)$ by the inextensibility condition, $(\frac{\partial x_2}{\partial s})^2+(\frac{\partial y_2}{\partial s})^2 = 1$, and thus
\begin{subequations}
    \begin{align}
        & x_2(s,t) = \int_0^{s} \cos(\phi(\sigma,t))d\sigma, \\
        & y_2(s,t) = \int_0^{s} \sin(\phi(\sigma,t))d\sigma.
    \end{align}
\end{subequations}

\subsection{Velocities and Accelerations}
Accelerations are indispensable for computing hydrodynamic reactive forces, and thus we derive expressions for the accelerations of the fish's head and body in addition to the velocity expressions. Denote the rotation matrices of $x_1y_1z_1$ and $x_2y_2z_2$ as $\bm{R}_{G_1}$ and $\bm{R}_{J}$. Denote the unit vectors of the inertial frame as $\{\hat{I},\hat{J},\hat{K}\}$ and the unit vectors of the head and joint frame as $\{\hat{i}_1,\hat{j}_1,\hat{k}_1\}$ and $\{\hat{i}_2,\hat{j}_2,\hat{k}_2\}$. We use the rotation matrices to express all unit vectors in the inertial frame. 

The velocity and acceleration of the head's COM can be written in the inertial frame as $\bm{v}_{G_1} = \dot{X}_{G_1}\hat{I}+\dot{Y}_{G_1}\hat{J}$ and $\bm{a}_{G_1} = \ddot{X}_{G_1}\hat{I}+\ddot{Y}_{G_1}\hat{J}$. The head's angular velocity $\bm{\omega}_{G_1}$ and angular acceleration $\bm{\alpha}_{G_1}$ are $\bm{\omega}_{G_1} = \dot{\theta}_{G_1}\hat{K}$ and $\bm{\alpha}_{G_1} = \ddot{\theta}_{G_1}\hat{K}$, and the velocity of point $J$ is $\bm{v}_J = \bm{v}_{G_1}+\bm{\omega}_{G_1} \times (-L_{G_1J} \hat{i}_1)$, where $L_{G_1J}$ denotes the distance from the head COM to the joint. The acceleration of point $J$ is $\bm{a}_J = \bm{a}_{G_1}+\bm{\alpha}_{G_1} \times (-L_{G_1J}\hat{i}_1)+\bm{\omega_{G_1}} \times (\bm{\omega_{G_1}} \times (-L_{G_1J}\hat{i}_1))$, and thus the velocity of a point on the fish's body with Lagrangian coordinate $s$ is $\bm{v}_{s} = \bm{v}_J+(\bm{v}_{s})_{x_2y_2z_2}+\bm{\omega}_{J} \times \bm{r}_{s/J}$, where $(\bm{v}_{s})_{x_2y_2z_2} = \frac{\partial x_2}{\partial t}\hat{i}_2+\frac{\partial y_2}{\partial t}\hat{j}_2$ is the velocity of the point as seen from $x_2y_2z_2$, and $\bm{r}_{s/J} = x_2\hat{i}+y_2\hat{j}$ denotes the relative position of the point with Lagrangian coordinate $s$ to the joint. The acceleration of a point on the fish's body is $\bm{a}_s  = \bm{a}_J+(\bm{a}_{s})_{x_2y_2z_2}+\bm{\alpha}_{G_1} \times \bm{r}_{s/J}
+\bm{\omega}_{G_1} \times (\bm{\omega}_{G_1} \times \bm{r}_{s/J})+2\bm{\omega}_{G_1} \times (\bm{v}_s)_{x_2y_2z_2}$, where $(\bm{a}_{s})_{x_2y_2z_2} = \frac{\partial ^2 x_2}{\partial s^2} \hat{i}_2+\frac{\partial ^2 y_2}{\partial s^2}\hat{j}_2$ is the acceleration of the point as seen from $x_2y_2z_2$.

\section{Energy Functionals and Virtual Work of Applied Forces}
\label{sec: Hamilton's principle}

We derive the equations of motion of the fish based on Hamilton's principle (a.k.a. Hamilton's action principle), which states that
\begin{align}
    \int_{t_0}^{t_f} (\delta \mathcal{L} + \delta W) dt = 0
    \label{eq: Hamilton's principle}
\end{align}
where $\mathcal{L} = T-V$ is the Lagrangian of the system ($T$ and $V$ stand for kinetic and potential energies), $\delta W$ stands for the virtual work done by forces that are not accounted for by potential energies, and the $\delta$ in front of $\mathcal{L}$ is the variational operator.

\subsection{Kinetic and Potential Energies}
The kinetic energy of the rigid head, $T_\text{head}$, is
\begin{align}
    T_\text{head} & = \frac{1}{2}m_1v_{G_1}^2+
                      \frac{1}{2}I_{G_1} \omega_{G_1}^2
\end{align}
where $m_1$ is the mass of the head and $I_{G_1}$ is the moment of inertia of the head about its COM along $z_1$. The kinetic energy of the fish's body is
\begin{align}
    T_\text{body} & = \int_0^{L_2} \frac{1}{2}\rho(s) v_{s}^2ds
\end{align}
where $\rho(s)$ is the linear density distribution of the fish body and $L_2$ is the length of the fish body. Then the total kinetic energy is $T = T_\text{head}+T_\text{body}$.

Since we focus on planar motions, only the body strain energy from elastic deformations contributes to the total potential energy; the potential energy can be written as
\begin{align}
    V
    & =
    \int_0^{L_2} \frac{1}{2}E(s)I(s)(\frac{\partial \phi}{\partial s})^2ds
\end{align}
where $E(s)$ is the Young's modulus of elasticity and $I(s)$ is the second moment of area of the body's cross-section (i.e., $EI$ is the bending rigidity of the fish's body), both of which can vary along the body as functions of the Lagrangian coordinate $s$.

\subsection{Joint Motor Torque}
\thispagestyle{empty}
Denote the motor torque as $\tau_m$; then the virtual work done by the motor torque and its reaction is
\begin{align}
    \delta W_m
    & = 
    \tau_m \hat{k}_2 \cdot (\overrightarrow{\delta\theta}_{G_1}+
    \overrightarrow{\delta\phi}|_{s=0}
    )
    -
    \tau_m \hat{k}_1 \cdot \overrightarrow{\delta\theta}_{G_1} \nonumber \\
    & = 
    \tau_m \hat{k}_2 \cdot \overrightarrow{\delta\phi}|_{s=0}
    = \tau_m \delta\phi|_{s=0}.
\end{align}

\subsection{Water Reactive Force and Moment on the Head}
Since the head is modeled as a rigid body, the water reactive force and moment on the head can be computed by the added mass matrix, $\bm{M}_a^\text{head}$:
\begin{align}
    \begin{bmatrix}
        \bm{F}_{r}^\text{head} \\
        \tau_{r}^\text{head}
    \end{bmatrix}
    = 
    -
    \begin{bmatrix}
        \bm{R}_{G_1} & 0 \\
        0            & 1
    \end{bmatrix}   
    \bm{M}_a^\text{head}
    \begin{bmatrix}
    \bm{R}_{G_1}^{\top}\bm{a}_{G_1} \\
    \alpha_{G_1}
    \end{bmatrix}
\end{align}
where the head's geometry is approximate as an ellipsoid with added mass matrix obtained from \cite{newman2018}. Then the virtual work is given by
\begin{align}
    \delta W_{r}^{\text{head}}
    = 
    \bm{F}_{r}^\text{head} \cdot \delta \bm{r}_{G_1}
    +\tau_{r}^\text{head}\delta\theta_{G_1}.
\end{align}

\subsection{Water Reactive Forces on the Body}
Lighthill’s large-amplitude elongated-body theory provides an integral momentum balance for evaluating the total reactive hydrodynamic force on a prescribed body motion \cite{lighthill1971}. Here, because the goal is to model the full robotic-fish dynamics, we use the corresponding differential momentum balance to obtain the distributed hydrodynamic load along the body.

\begin{figure}[!htb]
    \centering
    \includegraphics[width=0.95\linewidth,trim=0 0 0 15,clip]{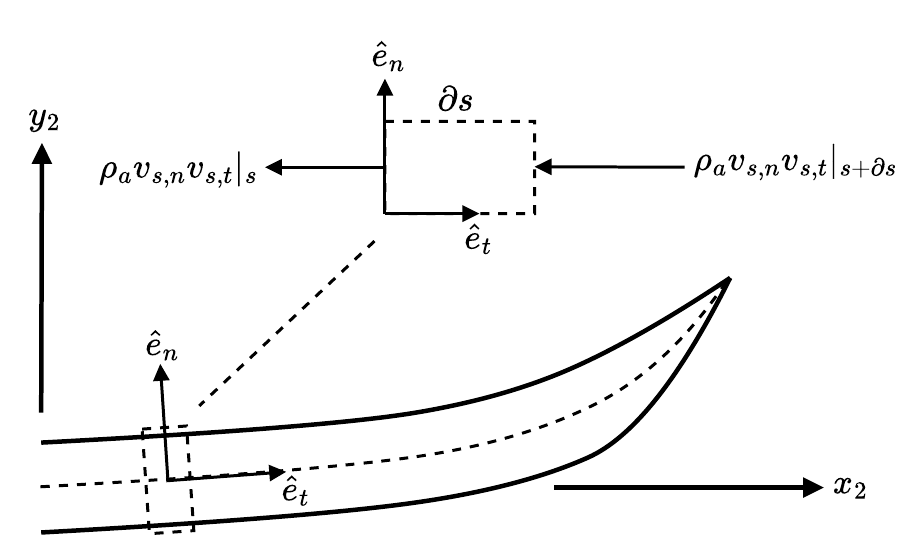}
    \caption{Differential control volume and its momentum flux of the water added mass around the fish's body.}
    \label{fig: CV}
\end{figure}

Fig. \ref{fig: CV} shows a differential control volume of the water added mass around the fish's body. The normal and tangential unit vectors can be written as
\begin{subequations}
    \begin{align}
            & \hat{e}_t = \frac{\partial x_2}{\partial s}\hat{i}_2
                          +\frac{\partial y_2}{\partial s}\hat{j}_2, \\
            & \hat{e}_n = -\frac{\partial y_2}{\partial s}\hat{i}_2
                          +\frac{\partial x_2}{\partial s}\hat{j}_2.
    \end{align}
\end{subequations}

According to Lighthill's large-amplitude elongated body theory, only the momentum in the normal direction (i.e., direction parallel to $\hat{e}_n$) is the dominant momentum in determining the water reactive forces. Denote $\bm{p}_n$ as the normal directional momentum per unit length; then
\begin{align}
    \bm{p}_n = \rho_a v_{s,n}\hat{e}_n
    \label{eq: momentum}
\end{align}
where $v_{s,n} = \bm{v}_s \cdot \hat{e}_n$ is the normal velocity component of the point on the fish's body with Lagrangian coordinate $s$. Then the momentum balance of the differential control volume is
\begin{align}
    \frac{d}{dt}(\bm{p}_n) = \underbrace{\frac{\partial}{\partial s}
                             (\rho_av_{s,n}v_{s,t})\hat{e}_n}
                             _\text{convective flux}
                             -\underbrace{\frac{\partial}{\partial s}
                             (\frac{1}{2}\rho_a v_{s,n}^2)\hat{e}_t}
                             _\text{dynamic pressure}
                             -\underbrace{\bm{f}_r}_\text{reactive force}
    \label{eq: momentum balance}
\end{align}
where $\rho_a = \frac{1}{4}\pi\rho_\text{water}h^2$ is the added mass per unit length ($h$ is the depth of the cross-sectional area), $\bm{f}_r$ is the water reactive force per unit length that the control volume applies to the fish body, and taking the time derivative of Eq.~\eqref{eq: momentum} yields
\begin{align}
    \frac{d}{dt}(\bm{p}_n) = \rho_a \frac{\partial v_{s,n}}{\partial t}
                             \hat{e}_n
                             +\rho_a v_{s,n}
                             (\bm{\omega}_{G_1}+\frac{\partial \phi}{\partial t} \hat{k}_2) \times \hat{e}_n.
    \label{eq: total derivative}
\end{align}
By equating Eq. (\ref{eq: momentum balance}) and Eq. (\ref{eq: total derivative}), we can solve for the reactive force per unit length, $\bm{f}_r$. Then the virtual work of water reactive forces on the fish's body is
\begin{align}
    \delta W_r^\text{body} = \int_0^{L_2} \bm{f}_r \cdot 
                             \delta \bm{r}_{s}ds.
\end{align}

\subsection{Water Drag Forces}
Since the fish moves at relatively low speeds, the drag forces on the fish's head and body can be modeled as linear damping, and the forces and moments can be written as
\begin{align}
    \begin{bmatrix}
        \bm{F}_{d}^\text{head} \\
        \tau_{d}^\text{head}
    \end{bmatrix}
    = 
    -
    \begin{bmatrix}
        \bm{R}_{G_1} & 0 \\
        0            & 1
    \end{bmatrix}   
    \bm{D}^\text{head}
    \begin{bmatrix}
    \bm{R}_{G_1}^{\top}\bm{v}_{G_1} \\
    \omega_{G_1}
    \end{bmatrix}
\end{align}
and
\begin{align}
    \bm{f}_d = -c_d\bm{v}_{s}
\end{align}
where $\bm{D}^\text{head}$ is the drag matrix of the rigid head and $c_d$ is the drag coefficient along the elastic body. The virtual work of drag forces is then
\begin{align}
    \delta W_d & = \bm{F}_{d}^\text{head} \cdot \delta \bm{r}_{G_1}
                   +\tau_{d}^\text{head}\delta\theta_{G_1}+
                   \int_0^{L_2} \bm{f}_d
                   \cdot \delta \bm{r}_s d{s}.
\end{align}

\section{Ritz Series and Computational Methods}
\label{sec: Ritz and computations}

Directly utilizing the derivations in Section \ref{sec: Hamilton's principle} leads to integro-partial differential equations (integro-PDEs) that pose challenges to obtaining numerical solutions. To obtain equations of motion that can be readily solved numerically, we implement the Ritz series method, high-precision Gauss-Legendre quadrature, and forward-mode automatic differentiation (AD) \cite{RevelsLubinPapamarkou2016} to obtain the projected weak form written as
\begin{align}
    \bm{M}\ddot{\bm{q}}^\text{c} = \bm{F}
    \label{eq: projected weak form}
\end{align}
where $\bm{M}$ is the mass matrix, $\bm{F}$ is the excitation vector, and $\bm{q}^\text{c}$ is the vector containing all generalized coordinates of the robotic fish (i.e., the vector representing the full configuration space of the robotic fish given the Ritz series). Then we can compute the instantaneous generalized accelerations at any instant by solving the linear system represented as Eq. (\ref{eq: projected weak form}) using direct methods. The system trajectory can then be obtained using numerical integrators for solving ordinary differential equations (ODEs).

\subsection{Ritz Series Approximations}
\thispagestyle{empty}
We approximate the displacement field represented by the tangential angle $\phi(s,t)$ as
\begin{align}
    \phi(s,t) = \sum_{n=1}^N q_n(t)\psi_n(s)
\end{align}
where $\psi_n(s)$ are the Ritz basis functions (shape functions), $q_n(t)$ are the Ritz coefficients (generalized coordinates), and $N$ is the total number of basis functions. It can be proven that as $N$ increases, this approximation would yield a representation that converges to the true response \cite{ginsberg2008}.

Since the Ritz basis functions set the spatial dependencies, the energy expressions are reduced from functionals of the displacement field (and its spatial and temporal derivatives) to functions of generalized coordinates (and their time derivatives). Specifically, the body coordinate $x_2$ can be written as
\begin{align}
    x_2(s,t) & = \int_0^s \cos\left(\sum_{n=1}^N 
                 q_n(t)\psi_n(\sigma)\right) 
                 d\sigma  \nonumber \\
             & = x_2(s,q_1,\ldots,q_N) \label{eq: x2 inner integral},
\end{align}
and its temporal derivative, obtained using the chain rule, is
\begin{align}
    \frac{\partial x_2}{\partial t}
             & = \int_0^s
                 -\left(
                 \sum_{n=1}^N \dot{q}_n(t)\psi_n(\sigma)
                 \right)
                 \sin\left(\sum_{n=1}^N 
                 q_n(t)\psi_n(\sigma)\right) 
                 d\sigma  \nonumber \\
             & = \dot{x}_2(s,q_1,\ldots,q_N,\dot{q}_1,\ldots,\dot{q}_N)
                 \label{eq: x2D inner integral}
\end{align}
where the dot above $x_2$ denotes a temporal derivative in a continuous system. The spatial derivative of $x_2$ can be written as
\begin{align}
    \frac{\partial x_2}{\partial s}
            & = \int_0^s
                -\left(
                \sum_{n=1}^N q_n(t)\psi_n'(\sigma)
                \right)
                \sin\left(\sum_{n=1}^N 
                q_n(t)\psi_n(\sigma)\right) 
                d\sigma  \nonumber \\
            & = x_2'(s,q_1,\ldots,q_N) \label{eq: x2P inner integral}
\end{align}
where the prime symbol near $x_2$ and $\psi_n$ denotes a spatial derivative in a continuous system. The coordinate $y_2$ and its spatial and temporal derivatives can be obtained similarly as functions of $s$, $q_n$'s, and time derivatives of $q_n$'s. Therefore, the energy functionals, which are integrals of $x_2$, $y_2$, and their spatial and temporal derivatives over the total length of the fish body, can then be represented as functions of $q_n$'s and their time derivatives.

Note that the integrals from Eq.~(\ref{eq: x2 inner integral}) to Eq.~(\ref{eq: x2P inner integral}) cannot be evaluated analytically for most basis function choices, and thus we numerically evaluate all the integrals using high-precision Gauss-Legendre quadrature. That is, we set up $K$ Gauss-Legendre nodes along the fish body, and at each node, we evaluate $x_2$, $y_2$, and their derivatives at this node (these are the inner integrals); then we compute the kinematics (i.e., velocities, accelerations, and curvatures) at the $K$ nodes using the inner integrals; finally, we compute the total kinetic and potential energies by the weighted Gauss-Legendre sum over the $K$ nodes (these are the outer integrals).

\subsection{Selections and Normalizations of Ritz Basis Functions}
Under this Ritz series representation, a rigid-body mode, $\psi_1(s) = 1$ (i.e., constant-angle mode), must be included in the basis functions to represent the rotation of the fish body about the joint.

Other basis functions of the Ritz series need to be linearly independent and satisfy the geometric boundary condition at the joint, which is $\phi(0,t) = 0$. Therefore, we start with monomial basis functions in addition to the rigid-body mode to capture deformations of the fish body; that is:
\begin{align}
    \hat{\psi}_n(s) = s^{n-1}, ~ n = 1,2,3,\ldots
\end{align}
To mitigate the ill-conditioning of the mass matrix and the numerical stiffness of the resulting ODEs, we normalize the basis functions about the density distribution of the fish body. Specifically, we define the $\rho$-norm of a basis function as $\| \hat{\psi}_n \|_\rho = \sqrt{\langle \hat{\psi}_n,\hat{\psi}_n \rangle_\rho}$, where $\langle \hat{\psi}_n,\hat{\psi}_n \rangle_\rho$ is the weighted inner product: $\langle \hat{\psi}_n,\hat{\psi}_n \rangle_\rho = \int_0^{L_2} \rho \hat{\psi}_n^2 ds$. Then the basis functions to use are
\begin{align}
    \psi_n = 
    \begin{cases}
        1 & \text{if} ~ n = 1 \\
        \frac{\hat{\psi}_n}{\|\hat{\psi}_n\|_\rho} & \text{if} ~ n > 1
    \end{cases}.
\end{align}

\subsection{The Projected Weak Form}
The Ritz series and Gauss-Legendre quadrature reduce Hamilton's principle for the system to Lagrange's equations. Specifically, let $\bm{q}^\text{c} = \begin{bmatrix} X_{G_1} & Y_{G_1} & \theta_{G_1} & q_1 & 
\ldots & q_N\end{bmatrix}^{\top}$, and let $q_j^\text{c}$ denote the $j$-th element in $\bm{q}^\text{c}$. With the displacement field written in Ritz series, the variation of the kinetic energy can be written as
\begin{align}
    \delta T(\bm{q}^\text{c},\dot{\bm{q}}^\text{c}) 
        = \sum_{j=1}^{N_q}
          (\frac{\partial T}{\partial q_j^\text{c}}\delta q_j^\text{c}+
          \frac{\partial T}{\partial \dot{q}_j^\text{c}}\delta \dot{q}_j^\text{c}).
\end{align}
Substituting it into the integration in Eq. (\ref{eq: Hamilton's principle}) gives
\begin{align}
    \int_{t_0}^{t_f} \delta T dt
    =
    \sum_{j=1}^{N_q} \int_{t_0}^{t_f}
    (\frac{\partial T}{\partial q_j^\text{c}}\delta q_j^\text{c}+
    \frac{\partial T}{\partial \dot{q}_j^\text{c}}\delta \dot{q}_j^\text{c}) dt.
\end{align}
Standard variational procedures of integration by parts and using the chain rule yield
% Integration by parts converts the time derivative $\delta \dot{q}_j^\text{c}$ into $\delta q_j^\text{c}$:
% \begin{align}
%     \int_{t_0}^{t_f} \delta T dt
%     & =
%     \sum_{j=1}^{N_q} \int_{t_0}^{t_f}
%     (\frac{\partial T}{\partial q_j^\text{c}}\delta q_j^\text{c}-
%     \frac{d}{dt}(\frac{\partial T}{\partial \dot{q}_j^\text{c}})\delta q_j^\text{c}) dt
%     \nonumber \\
%     & \qquad \qquad \qquad \qquad ~
%     +\underbrace{\frac{\partial T}
%     {\partial \dot{q}_j^\text{c}}\delta q_j^\text{c} |_{t_0}^{t_f}}_0.
% \end{align}
% The chain rule gives
% \begin{align}
%     \frac{d}{dt}(\frac{\partial T}{\partial \dot{q}_j^\text{c}})
%     =
%     \sum_{k=1}^{N_q} \left(
%     \frac{\partial^2 T}
%     {\partial\dot{q}_j^\text{c}\partial\dot{q}^\text{c}_k}
%     \ddot{q}^\text{c}_k+
%     \frac{\partial^2 T}{\partial\dot{q}_j^\text{c}\partial{q}^\text{c}_k}
%     \dot{q}^\text{c}_k
%     \right).
% \end{align}
% Therefore,
\begin{align}
    \int_{t_0}^{t_f} \delta T dt
    & =
    \int_{t_0}^{t_f}
    \left(
    \nabla_{\bm{q}^\text{c}}T
    -\frac{\partial^2 T}{\partial \dot{\bm{q}}^{\text{c}^2}}
    \ddot{\bm{q}}^\text{c}
    -\frac{\partial^2 T}{\partial \dot{\bm{q}}^\text{c} \partial 
    \bm{q}^\text{c}}\dot{\bm{q}}^\text{c}
    \right)^{\top} \delta \bm{q}^\text{c} dt.
    \label{eq: kinetic energy term}
\end{align}
Similarly,
\begin{align}
    \int_{t_0}^{t_f} \delta V dt
    =
    \int_{t_0}^{t_f} \left( \nabla_{\bm{q}^\text{c}}V \right)^{\top} \delta
    \bm{q}^\text{c} dt.
    \label{eq: potential energy term}
\end{align}

We proceed to derive the projections of the forcing terms (i.e., forces that contribute to the virtual work in Eq.~\eqref{eq: Hamilton's principle}) onto the generalized coordinates. The projections are the generalized forces given the Ritz series. Since
\begin{align}
    \delta \phi |_{s=0} = \sum_{n=1}^N \frac{\partial \phi_n}{\partial q_n}
                          \delta q_n |_{s=0}
                        = \frac{\partial \phi_1}{\partial q_1}\delta q_1
                        = \delta q_1,
\end{align}
then the generalized forces of the joint motor torque are
$
    \bm{Q}_m = \begin{bmatrix}
                    0 & 0 & 0 & \tau_m & 0 & \ldots & 0
               \end{bmatrix}^{\top}
    \label{eq: Q_m}
$. Since the generalized coordinates of the head are chosen from fixed references, the generalized forces of the water reactive forces on the head are
$
    \bm{Q}_r^\text{head}
    =
    \begin{bmatrix}
        -\bm{M}_a^\text{head,eff} & 0_{3\times N} \\
        0_{N\times 3} & 0_{N\times N}
    \end{bmatrix}\ddot{\bm{q}}^\text{c}
    \label{eq: Q_r_head}
$
, where
$
    \bm{M}_a^\text{head,eff}
    =
    \begin{bmatrix}
        \bm{R}_{G_1} & 0 \\
        0            & 1
    \end{bmatrix}   
    \bm{M}_a^\text{head}
    \begin{bmatrix}
        \bm{R}_{G_1}^{\top} & 0 \\
        0 & 1
    \end{bmatrix}
$. Similarly, the generalized forces of water drag on the head can be written as
$
    \bm{Q}_d^\text{head}
    =
    \begin{bmatrix}
        -\bm{D}^\text{head,eff} & 0_{3\times N} \\
        0_{N\times 3} & 0_{N\times N}
    \end{bmatrix}\dot{\bm{q}}^\text{c}
    \label{eq: Q_d_head}
$
, where
$
    \bm{D}^\text{head,eff}
    =
    \begin{bmatrix}
        \bm{R}_{G_1} & 0 \\
        0            & 1
    \end{bmatrix}   
    \bm{D}^\text{head}
    \begin{bmatrix}
        \bm{R}_{G_1}^{\top} & 0\\
        0 & 1
    \end{bmatrix}
$.

Denote the Jacobian matrix at a Gauss-Legendre node as $\bm{J}_s = \frac{\partial \bm{r}_s}{\partial \bm{q}^\text{c}}$, then the virtual work done by the water reactive forces on the body is
\begin{align}
    \delta W_r^\text{body}
    & =
    \int_0^{L_2}
    \bm{f}_r^{\top} \delta \bm{r}_s ds 
    = 
    \int_0^{L_2}
    \bm{f}_r^{\top} \bm{J}_s \delta \bm{q}^\text{c} ds \nonumber \\
    & = 
    \left(\int_0^{L_2}
    \bm{J}_s^{\top} \bm{f}_r ds \right)^{\top} \delta \bm{q}^\text{c}
    \label{eq: Q_r_body}
\end{align}
Similarly, the virtual work done by the water drag on the body is
\begin{align}
    \delta W_d^\text{body}
    & =
    \int_0^{L_2}
    \bm{f}_d^{\top} \delta \bm{r}_s ds 
    = 
    \left(\int_0^{L_2}
    \bm{J}_s^{\top} \bm{f}_d ds \right)^{\top} \delta \bm{q}^\text{c}
    \label{eq: Q_d_body}
\end{align}

Substituting Eqs. (\ref{eq: kinetic energy term}), (\ref{eq: potential energy term}), $\bm{Q}_m$, $\bm{Q}_r^\text{head}$, $\bm{Q}_d^\text{head}$, and Eqs. (\ref{eq: Q_r_body}) and (\ref{eq: Q_d_body}) into Eq. (\ref{eq: Hamilton's principle}), the fundamental lemma of the calculus of variations yields the ODEs governing the motions of the robotic fish. Move all terms containing generalized accelerations to the left side of the ODEs to incorporate them into the mass matrix, and then we obtain the final projected weak form in Eq. (\ref{eq: projected weak form}), where
\begin{align}
    \bm{M}
    =
    \frac{\partial^2 T}{\partial \dot{\bm{q}}^{\text{c}^2}}
    + \begin{bmatrix}
        \bm{M}_a^\text{head,eff} & 0 \\
        0 & 0
    \end{bmatrix}
    - \frac{\partial}{\partial \ddot{\bm{q}}^\text{c}}
    \int_0^{L_2}\bm{J}_s^{\top} \bm{f}_r ds
\end{align}
and
\begin{align}
    \bm{F}
    & =
    \nabla_{\bm{q}^\text{c}}T
    -\frac{\partial^2 T}{\partial \dot{\bm{q}}^\text{c} \partial 
    \bm{q}^\text{c}}\dot{\bm{q}}^\text{c}
    -\nabla_{\bm{q}^\text{c}}V
    +\bm{Q}_m
    +\bm{Q}_d^\text{head} \nonumber \\
    & \qquad \qquad
    + \int_0^{L_2}\bm{J}_s^{\top} \bm{f}_r ds \Big|_{\ddot{\bm{q}}^\text{c}=0}
    +\int_0^{L_2}\bm{J}_s^{\top} \bm{f}_d ds
\end{align}
To optimize simulation code performance, local derivatives at each outer Gauss-Legendre node are computed symbolically and then assembled inside the ODE function definition. The resulting ODEs can then be solved using numerical integrators that are accessible in most scientific computing languages. 

\subsection{Algorithm Implementation}

We implement the numerical simulations in the Julia programming language. Since the ODE system is highly stiff with strong couplings, implicit methods are needed. We use two integrators from the DifferentialEquations.jl package -- 1) the TRBDF2 method (an efficient solver that combines the trapezoidal rule with the second-order backward difference formula) with adaptive time steps and 2) the Rodas5P method (a higher-order and more accurate solver that uses the Rosenbrock method) with a fixed time step of 0.001 \si{\second}  \cite{rackauckas2017}. Our Julia code runs on a desktop with an AMD Ryzen 9 9950X3D2 Dual Edition 4.3 GHz processor and 32 GB DDR5 RAM.

\section{Results}
\thispagestyle{empty}
In this section, we validate the dynamics model via numerical convergence and hardware experiments. Building on the validated dynamics model, we then utilize the differentiability of the model to perform a gradient-based design optimization over the fish body's stiffness distribution to balance propulsion speed and efficiency.

% develop a full pipeline for design optimizations, speed and efficiency analysis, and feedforward-feedback control, which demonstrates the capabilities of the differentiable dynamics model and fast simulator.

\subsection{Ritz Series Convergence}
We first validate the dynamics model by testing the convergence of the Ritz series and, at the same time, choose the number of Ritz basis functions that are approximately complete. We define the Young's modulus distribution using the polynomials $E(s) = (0.35-0.7s^3) \times 10^6$ \si{\pascal}. That is, we set the baseline Young's modulus to 350 kPa, as this is the Young's modulus of typical bio-based materials \cite{cui2022}; we also use monotonically decreasing functions to reduce stiffness near the tail, as this specific type of distribution has been shown to provide efficient propulsion in robotic fish designs \cite{wang2021effect}.
% Matching our hardware setup, the Young's modulus distribution of the fish body is fixed throughout the validation experiments and is defined via the polynomials $E(s) = (0.35-0.7s^3) \times 10^6$ \si{\pascal}. 

\begin{figure}[!htb]
    \centering
    \includegraphics[width=0.95\linewidth,trim=5 0 0 0,clip]{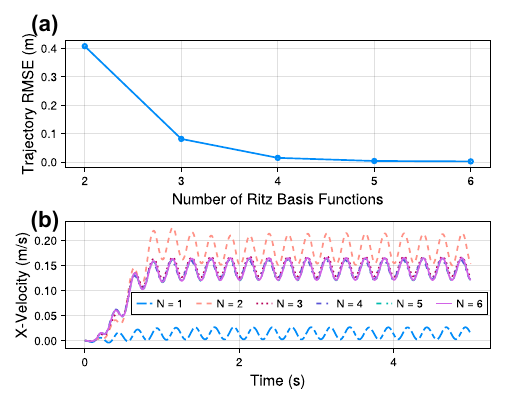}
    \caption{Convergence analysis of the Ritz series. (a) Trajectory root-mean-square error as the number of Ritz basis functions increases. (b) Time evolution of $X$-directional velocities with an increasing number of Ritz basis functions.}
    \label{fig: convergence}
\end{figure}

For convergence analysis, we first compare 5-second trajectories (where the robotic fish has fully entered steady state) simulated under an increasing number of Ritz basis functions. We use PD control that enforces the motor angle to oscillate at a frequency of 2 \si{\hertz} and an amplitude of 25\si{\degree}. Define the root-mean-square error (RMSE) between two trajectories of the same duration as 
$
    \text{RMSE}(\bm{X}_1,\bm{X}_2) 
    = 
    \frac{\|\bm{X}_2-\bm{X}_1\|_F}{\sqrt{N_\text{traj}}}
$,
where $\bm{X}_1, \bm{X}_2 \in \mathbb{R}^{2 \times N_\text{traj}}$ are the evenly sampled trajectory matrices of the fish head's COM position, $\| \cdot \|_F$ is the Frobenius norm, and $N_\text{traj}$ is the duration of the sampled trajectories. Fig.~\ref{fig: convergence}(a) shows the convergence of trajectory RMSE to 0 \si{\meter} as the number of Ritz basis functions increases, where the trajectory RMSE with $N$ Ritz basis functions is defined as the RMSE of the trajectories with $N$ and $N-1$ basis functions.

% (c)-(d) Trajectories of the robotic fish swimming at \SI{2}{\hertz} and \SI{15}{\degree} (c) and \SI{25}{\degree} (d) joint motor oscillation angles. A camera (Hero 13 Black, GoPro) was positioned overhead to capture the videos, with positions shown every eight seconds. The scale bar is 0.25 BL (10 cm). (e) Comparison of simulated and hardware steady-state speeds at 15\si{\degree} and 25\si{\degree} joint motor oscillation angles.

In addition to the head COM position, we also investigate how the velocity of head COM in the $X$ direction evolves with time, and more importantly whether this evolution converges as the number of Ritz basis functions increases. As shown in Fig.~\ref{fig: convergence}(b), using only the rigid-body mode severely underestimates the steady-state speed, and adding a constant curvature will overestimate the steady-state speed. As $N$ reaches 3, the time evolution of the speed, including the steady-state speed, does not significantly change as $N$ increases, which again implies convergence.

Simulating a 5-second trajectory using 6 Ritz basis functions takes 1.15 seconds on average. Given the previous convergence analysis, using 6 basis functions balances performance and accuracy, with simulation speed exceeding the requirement of real-time simulations. The fast simulations not only represent a significant improvement compared to existing FSI-based methods, but is also necessary considering that we aim to pass dual numbers into the ODE functions and solvers to compute gradients by forward-mode automatic differentiation and perform iterative gradient-based optimizations. 

\subsection{Design of Robotic Fish and Experimental Validation}
\begin{figure}[!htb]
\setlength{\abovecaptionskip}{-0.1cm}
\setlength{\belowcaptionskip}{-1cm}
\centering
\includegraphics[width=3.49in]{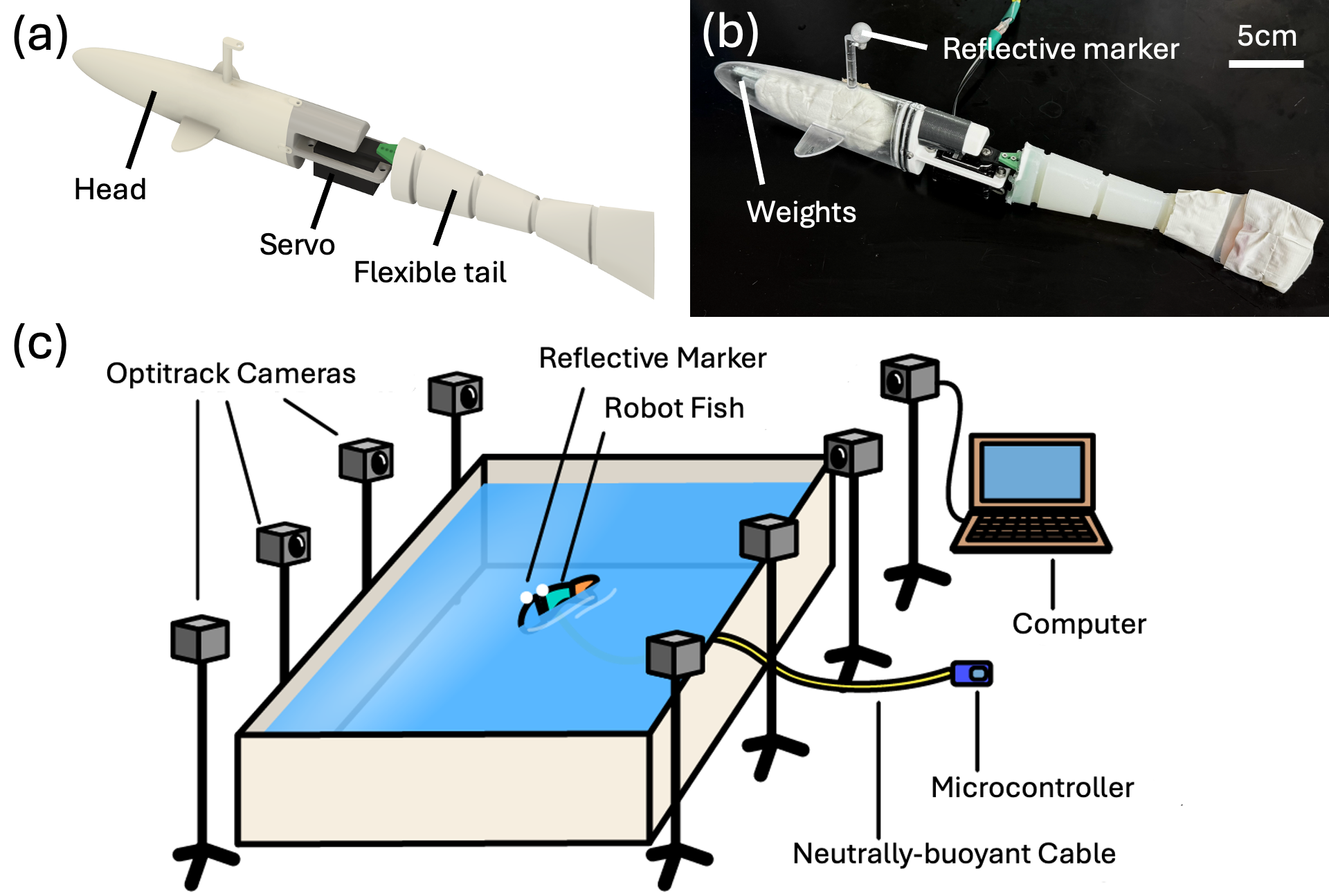}
\caption{
Hardware design and experimental setup. (a) Schematic of the robotic fish, consisting of a streamlined rigid head, a servo motor, and a flexible tail. (b) Prototype of the baseline robotic fish design. (c) Experimental setup for free-swimming tests. Eight OptiTrack motion-capture cameras were arranged around a \SI{5}{m}$\times$\SI{2}{m} pool, and a reflective marker mounted on the robot was used for position tracking. Motor commands were transmitted from the microcontroller through a neutrally buoyant cable. The microcontroller and motion-capture system communicated with the host computer via Wi-Fi and Ethernet, respectively.
} 
\label{fig:hardware}
% \vspace{-0.5cm}
\end{figure}

\thispagestyle{empty}
% written by Jiayi Jin
To further validate the dynamics model, we developed a robotic fish with the same rigid-head, motor-driven joint, and passive flexible-tail configuration used in simulation (Fig.~\ref{fig:hardware}(a)). The rigid head is a streamlined \SI{1.5}{mm}-thick shell that houses balancing weights and a reflective marker for motion tracking. A low-profile waterproof servo (DB778WP, Hitec) was placed where it is consistent with the model configuration.

In the dynamics model, the tail flexural rigidity is determined by the spatially varying Young’s modulus $E(s)$ and the second moment of area $I_{\text{body}}(s)$ associated with the local body geometry, where $E(s) = (0.35-0.7s^3) \times 10^6$ \si{\pascal}. To fabricate the tail as a monolithic structure with a uniform material modulus, the continuous stiffness distribution was reproduced using discrete reduced-cross-section grooves. At each discretization location $s_i$, the groove geometry was determined by matching the integrated flexural compliance of the physical tail to that of the numerical model over a local window of length $2\Delta s$:

\begin{small}
\begin{multline}
\int_{s_i-\Delta s}^{s_i+\Delta s}\frac{\mathrm{d}s}{E(s)I_\text{body}(s)}\\
=\frac{1}{E_\text{sil}}\left[ \int_{s_i-\Delta s}^{s_i-w_c/2}\frac{\mathrm{d}s}{I_\text{body}(s)} + \frac{w_c}{I_i^c} + \int_{s_i+w_c/2}^{s_i+\Delta s}\frac{\mathrm{d}s}{I_\text{body}(s)} \right],
\label{eqn:tail-design}
\end{multline} 
\end{small}
where $E_{\mathrm{sil}}$ is the silicone modulus, $\Delta s$ is the spacing between adjacent discretization locations, and $w_c$ and $I_i^c$ are the longitudinal width and second moment of area of the $i$-th grooved segment, respectively. In the present design, $\Delta s=40~\mathrm{mm}$ and $w_c=5~\mathrm{mm}$. The groove cross section was approximated as an ellipse, such that $I_i^c=\frac{\pi}{64}h_i^c\left(b_i^c\right)^3$, where $h_i^c$ and $b_i^c$ denote the dimensions parallel and perpendicular to the neutral axis, respectively. These dimensions were calculated from the required $I_i^c$ obtained from Eq. \eqref{eqn:tail-design} and were constrained by the local external body envelope and a minimum manufacturable feature size of \SI{1}{mm}. Among feasible solutions, larger normalized cross-sectional dimensions were preferred to improve fabrication robustness.

The prototype has a body length of \SI{40}{cm} and a mass of approximately \SI{306}{g} (Fig.~\ref{fig:hardware}(b)). The head was 3D printed with ultraviolet-curable resin (Clear Resin v5, Formlabs) using a desktop stereolithography printer (Form 4, Formlabs) for optimal waterproofing. The servo-mounting cap was printed with acrylonitrile butadiene styrene (ABS, Bambu Lab) using a desktop 3D printer (X1C, Bambu Lab) for impact resistance. The cap was fixed to the head with screws, and two silicone O-rings were inserted to ensure sealing. 
The flexible tail was fabricated as a single monolithic component through silicone casting in a negative mold produced from the final tail design. Dragon Skin 30 ($E_\text{sil}=593~\mathrm{kPa}$, Smooth-On Inc.) was selected to provide flexural stiffness comparable to that of biological fish \cite{wang2021effect}. 
Additional weights were placed inside the head to achieve near-neutral buoyancy and an approximately level posture at rest.

Free-swimming tests were conducted in a \SI{5}{m} by \SI{2}{m} pool using an OptiTrack motion-tracking system (Fig.~\ref{fig:hardware}(c)). Swimming trajectories were collected over multiple driving frequencies and motor amplitudes, with three trials per condition.
The robot was controlled by a microcontroller board (XIAO ESP32S3, Seeed Studio) through Wi-Fi. Power was supplied by a two-cell lithium-polymer battery with a maximum charged voltage of \SI{8.4}{V} (2S 650 mAh LiPo-Battery, Zeee). The servo voltage, current, and power were measured with a power sensor (INA219, Adafruit). The servo was connected through a neutrally buoyant tether cable (Fathom ROV Tether, Blue Robotics).

\begin{figure}[!htb]
    \centering
    \includegraphics[width=0.95\linewidth,trim=5 5 0 0,clip]{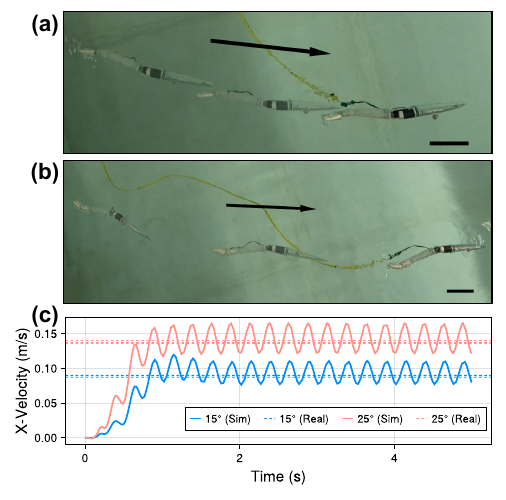}
    \caption{Experimental validation results. (a)-(b) Trajectories of the robotic fish swimming at \SI{2}{\hertz} and \SI{15}{\degree} (a) and \SI{25}{\degree} (b) joint motor oscillation angles. A camera (Hero 13 Black, GoPro) was positioned overhead to capture the videos, with positions shown every eight seconds. The scale bar is 0.25 BL (10 cm). (c) Comparison of simulated and hardware steady-state speeds at 15\si{\degree} and 25\si{\degree} joint motor oscillation angles.}
    \label{fig: experiment}
\end{figure}

Finally, using 6 Ritz basis functions, we compare the simulated steady-state speeds and the hardware's true steady-state speeds at different motor oscillation amplitudes. We let the motor oscillate at a frequency of \SI{2}{\hertz}, and perform three hardware experiments at \SI{15}{\degree}, \SI{20}{\degree}, and \SI{25}{\degree} oscillation amplitudes, respectively. We use the \SI{20}{\degree} experimental data to calibrate the drag coefficients of the dynamics model, and then use the \SI{15}{\degree} and \SI{25}{\degree} experimental data to validate the model. Snapshots of the fish's configuration during hardware experiments are shown in Fig.~\ref{fig: experiment}(a) and Fig.~\ref{fig: experiment}(b). As shown in Fig.~\ref{fig: experiment}(c), at both \SI{15}{\degree} and \SI{25}{\degree} oscillation amplitudes, the simulated steady-state speeds match the real steady-state speeds with high precision. This implies that our simulation provides a decent prediction of the steady state of the robotic fish system.

\subsection{Stiffness Distribution Optimizations}
\thispagestyle{empty}
To further demonstrate the model’s differentiability and capability for real-time optimization, we optimize the fish-body stiffness distribution using an objective that balances steady-state swimming speed and cost of transport (COT). For this robotic platform, we define COT specifically as the electrical energy consumed per unit of forward distance:
\begin{align}
    \text{COT} 
    = 
    \frac{\Delta X_{G_1}}{\int_{t_0}^{t_f}\left| P_m(t) \right| dt}
\end{align}
where $\Delta X_{G_1}$ is the forward displacement of the head's COM, and $P_m(t)$ is the power of the joint motor at time $t$. The cost function is then defined as 
\begin{align}
    J 
    = 
    (1-w_\text{speed})\overline{\text{COT}}
    - w_\text{speed}\overline{v}_{ss}
    \label{eq: cost}
\end{align}
where $w_\text{speed} \in [0,1]$ is the weight of the steady-state speed, $\overline{\text{COT}}$ is the normalized COT, and $\overline{v}_{ss}$ is the normalized steady-state speed.

Since existing research shows that the stiffness of a fish body decreases from head to tail for efficient propulsions \cite{wang2021effect,zhang2026fish}, we use sum-of-squares parameterization and describe the Young's modulus distribution of the fish body as 
$
    E(s) = p_{E,0}-p_{E,1}^2s-\frac{p_{E,2}^2}{2}s^2
           -\frac{p_{E,3}^2}{3}s^3-\frac{p_{E,4}^2}{4}s^4.
$
That is, the Young's modulus distribution is determined by the polynomial's coefficient vector $\bm{p}_E = [p_{E,0} \ \ldots \ p_{E,4}]^T$. We fix the joint motor oscillation amplitude to 25\si{\degree} and frequency to 2 \si{\hertz}. Then the cost in Eq.~\eqref{eq: cost} is only a function of the stiffness parameters, and the optimization problem of finding the optimal stiffness distribution can be formulated as
\begin{align}
    \bm{p}_E^* 
    = 
    \arg\min_{\bm{p}_E} J(\bm{p}_E) 
    \quad \text{s.t.} \ E(L_2) \geq 10^5 ~ \si{\pascal}.
    \label{eq: optimization}
\end{align}
The constraint is added to avoid unphysical Young's modulus that gives an overly-soft body. 

We solve the optimization problem presented in Eq.~\eqref{eq: optimization} by the limited-memory Broyden-Fletcher-Goldfarb-Shanno (L-BFGS) algorithm with backtracking line search, and the constraint is enforced by adding a barrier function term during the code implementation. We choose L-BFGS rather than gradient descent because L-BFGS is a second-order method that provides superlinear convergence, making it more suitable for the case where the time consumed by a single iteration is nontrivial. In each L-BFGS iteration, forward-mode automatic differentiation passes dual numbers into the forward simulation, obtaining the cost function value and its exact gradient. The inverse Hessian, whose multiplication with the gradient determines the search direction, is then approximated recursively using the gradient history.

\begin{figure}[!htb]
    \centering
    \includegraphics[width=1\linewidth,trim=0 5 5 0,clip]{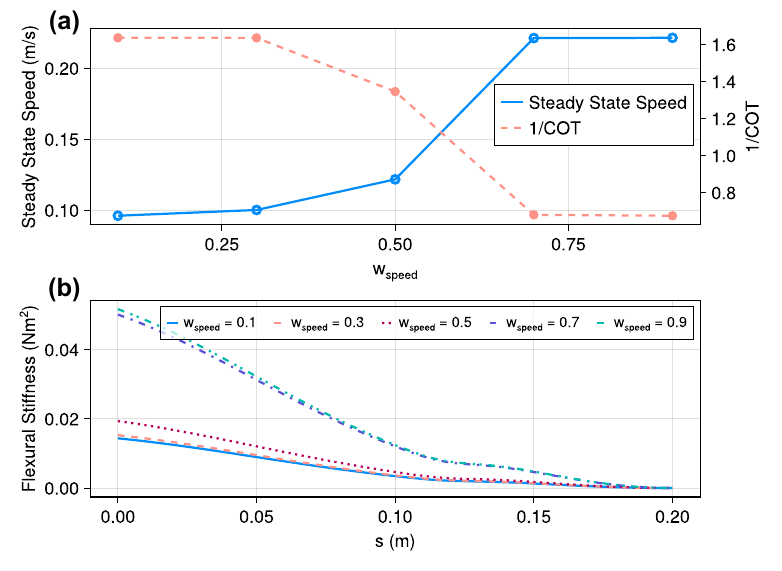}
    \caption{Body stiffness optimization results. (a) Steady-state speeds and efficiencies of optimized stiffness distributions, and (b) optimized flexural stiffness distributions along the fish body,  under different $w_\text{speed}$ values.}
    \label{fig: stiffness}
\end{figure}

We perform optimizations using $w_\text{speed}$ values of 0.1, 0.3, 0.5, 0.7, and 0.9. The steady-state speeds and the efficiencies (i.e., $\frac{1}{\text{COT}}$) resulted from the optimized Young's modulus distributions at different $w_\text{speed}$ values is shown in Fig.~\ref{fig: stiffness}(a). The resulting flexural stiffness (i.e., $EI$) distributions with different $w_\text{speed}$ values are shown in Fig.~\ref{fig: stiffness}(b).

The trend in Fig.~\ref{fig: stiffness}(a) matches the existing theory that speed and efficiency are antagonistic in fish locomotion. The plot also reveals the upper and lower bounds of the robotic fish's speed and efficiency from only optimizing its body stiffness distribution. Fig.~\ref{fig: stiffness}(b) shows that an overall softer body gives higher efficiency but lower speed, while an overall stiffer body gives higher speed but lower efficiency.

\section{Conclusion and Future Work}
\thispagestyle{empty}
To conclude, our work presents a physics-grounded differentiable model and fast simulations for the full-body dynamics of fish swimming. Our model captures the dynamics of the distributed elasticity of the fish body under large deformations, while also incorporating reactive and resistive hydrodynamic forces, offering a continuous and computationally tractable alternative to multi-segment or FSI-based approaches. Numerical and hardware experiments validate the model, and gradient-based optimizations over the body stiffness distribution demonstrate the differentiability of the simulation framework.

Our future work includes extending the model to 3D maneuvers and more complex hydrodynamics, implementing model-based control or reinforcement learning, performing co-optimizations by including body shape and controller parameters, and further model reductions so that real-time simulations are possible on resource-limited embedded systems.

% A preliminary parameter study reveals how actuation frequency and body stiffness affect speed and efficiency, and how body stiffness distribution can affect the accuracy and energy efficiency of closed-loop control at different target speeds.

% Our future work includes utilizing the automatically differentiable nature of our dynamics model to perform high-dimensional numerical optimizations over the full parameter space to find the optimal geometries and material properties for maximizing swimming speed and efficiency. Moreover, we are interested in further model reductions to enable real-time model-based controls with hardware implementations.

% \clearpage

\bibliographystyle{IEEEtran}
\bibliography{references}

\end{document}